\documentclass[trans]{IEEEtran}

\usepackage[colorlinks,urlcolor=blue,linkcolor=blue,citecolor=blue]{hyperref}
\usepackage{amsmath,amsfonts}
\usepackage{algorithmic}
\usepackage{algorithm}
\usepackage{array}
\usepackage{textcomp}
\usepackage{stfloats}
\usepackage{url}
\usepackage{verbatim}
\usepackage{graphicx}
\usepackage{cite}
\usepackage{booktabs}
\usepackage{multirow}
\usepackage{threeparttable}
\usepackage{dcolumn}
\usepackage{siunitx}
\usepackage{bm}
\usepackage[table]{xcolor}
\begin{document}

\title{FM-Planner: Foundation Model Guided Path Planning for Autonomous Drone Navigation}

\author{Jiaping Xiao, Cheng Wen Tsao, Yuhang Zhang and Mir Feroskhan,~\IEEEmembership{Member, IEEE}
\thanks{J. Xiao, C. W. Tsao, Y. Zhang, and M. Feroskhan are with the School of Mechanical and Aerospace Engineering, Nanyang Technological University, Singapore 639798, Singapore (e-mail: jiaping001@e.ntu.edu.sg; tsao0002@e.ntu.edu.sg; yuhang004@e.ntu.edu.sg;  mir.feroskhan@ntu.edu.sg).
\textit{(Corresponding author: Mir Feroskhan.)}}
}



\maketitle

\begin{abstract}

Path planning is a critical component in autonomous drone operations, enabling safe and efficient navigation through complex environments. Recent advances in foundation models, particularly large language models (LLMs) and vision-language models (VLMs), have opened new opportunities for enhanced perception and intelligent decision-making in robotics. However, their practical applicability and effectiveness in global path planning remain relatively unexplored. This paper proposes foundation model-guided path planners (FM-Planner) and presents a comprehensive benchmarking study and practical validation for drone path planning. Specifically, we first systematically evaluate eight representative LLM and VLM approaches using standardized simulation scenarios. To enable effective real-time navigation, we then design an integrated LLM-Vision planner that combines semantic reasoning with visual perception. Furthermore, we deploy and validate the proposed path planner through real-world experiments under multiple configurations. Our findings provide valuable insights into the strengths, limitations, and feasibility of deploying foundation models in real-world drone applications and providing  practical implementations in autonomous flight. Project site: \url{https://github.com/NTU-ICG/FM-Planner}.
\end{abstract}


\section{Introduction}

{\IEEEPARstart{A}{utonomous} navigation capabilities are essential for drones across diverse applications, including search and rescue operations \cite{xiao2023collaborative}, parcel delivery \cite{kornatowski2020downside}, infrastructure inspection \cite{liu2023industrial}, and environmental monitoring \cite{jia2024multitarget}. Effective path planning is a critical element of autonomous flight, ensuring safe and efficient navigation by generating trajectories that guide drones through dynamic, cluttered, and often unpredictable environments. Traditional approaches to drone path planning, such as the A* algorithm \cite{zhou2022uav} and Rapidly-exploring Random Trees (RRT) \cite{hu2021efficient}, have been widely adopted due to their simplicity but face challenges in dynamic or high-dimensional environments where adaptability and semantic awareness are required. Optimization-based methods, such as Fast-Planner \cite{zhou2019robust} and EGO-Planner \cite{zhou2021ego} address some of these limitations by explicitly formulating path planning as constrained optimization problems with desired objectives. Nevertheless, these methods typically rely on accurate models of the environment and robot dynamics, making them less flexible in unstructured or rapidly changing environments.

Recently, learning-based approaches have shown significant promise for their ability to handle complex path planning tasks without explicit modeling of dynamics and optimization \cite{xiaovision2025}. Imitation learning (IL) and deep reinforcement learning (DRL) have been applied to achieve agile navigation \cite{sun2022aggressive, xiaolearning2024}, perception-aware flight \cite{zhang2025learning}, championship-level drone racing \cite{kaufmann2023champion} and collaborative navigation \cite{chen2023multiagent, zhang2024efficient, xiao2024toward} and formation maneuvering \cite{yan2022deep, xiao2025resilient}. While these methods demonstrate strong performance within trained environments, they require large amounts of domain-specific data for each task and often struggle to generalize to unseen scenarios.

The emergence of foundation models, notably large language models (LLMs) \cite{achiam2023gpt} and vision-language models (VLMs) \cite{alayrac2022flamingo}, has significantly advanced the field of artificial intelligence by providing powerful tools for enhanced perception, reasoning, and decision-making across diverse applications, such as design \cite{lin2025pegpt, qin2025harnessing}, monitoring \cite{wang2024large} and disassembly \cite{xia2025leveraging}. Foundation models are pretrained on large-scale and varied datasets, enabling them to generalize effectively across different tasks with minimal fine-tuning. Recent robotics applications leveraging foundation models include robotic manipulation \cite{cui2022can}, navigation \cite{wang2024towards}, and human-robot interaction \cite{firoozi2023foundation}. Despite these promising results, the applicability, performance, and challenges of deploying foundation models for global path planning tasks remain largely unexplored for drones. Unlike previous learning-based methods focused on reactive control or environment awareness, foundation models offer the potential to perform high-level spatial reasoning and interpret multimodal prompts for trajectory generation. While existing works such as LEVIOSA \cite{aikins2024leviosa} convert natural language commands to drone waypoints, they are limited to local inference and lack systematic evaluation across varied scenarios or integration with real-time perception.

\begin{figure}[!tb]
    \centering
    \includegraphics[width=\linewidth]{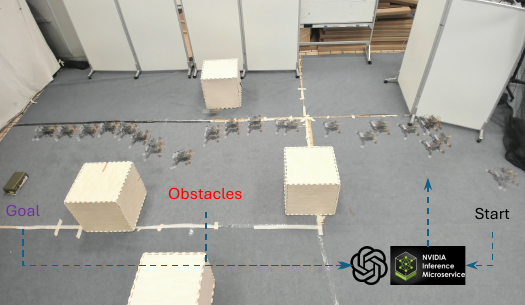}
    \caption{The real-world autonomous flight with a foundation model-guided path planner. The obstacle positions are recognized by a vision encoder and fed into LLM with prompts.}
    \label{fig:realflight}
\end{figure}
In this work, we propose a foundation model-guided path planning framework, FM-planner, tailored specifically for autonomous drone navigation. We conduct an extensive benchmarking study that systematically evaluates eight state-of-the-art large language models (LLMs) and five vision-language models (VLMs) across diverse simulated scenarios. Leveraging insights gained from this benchmark, we further fine-tune the top-performing LLM, Llama-3.1-8B-Instruct, using trajectories collected from representative drone flights. Subsequently, we integrate this fine-tuned LLM model with a general-purpose vision encoder (LLM-Vision) to enable real-time obstacle perception and path planning (see Fig.~\ref{fig:realflight}). The main contributions of this research are summarized as follows:
\begin{enumerate}
    \item We propose a foundation model-guided path planning framework designed specifically for autonomous drone navigation, integrating tailored prompt engineering and a multistage architecture.
    \item To our knowledge, this work is the first to comprehensively benchmark multiple foundation models for drone path planning, systematically assessing their performance across diverse simulation scenarios.
    \item We further deploy and validate the effectiveness of LLM-Vision-guided path planner through physical drone experiments in real-world flight scenarios.
\end{enumerate}

The rest of this article is organized as follows. The related work is discussed in Section II. Section III introduces the proposed foundation model guided path planners. The experiment settings and results are presented and discussed in Section IV. Section V concludes this article.

\section{Related Works}
\label{sec:relatedwork}

\subsection{Path Planning}
Path planning is a fundamental problem in robotics, involving the computation of optimal or feasible paths from an initial location to a destination, under constraints such as obstacle avoidance and efficient traversal. Traditional path planning methods can generally be classified into grid-based algorithms, like A* \cite{hart1968formal}, and sampling-based algorithms, such as Rapidly-exploring Random Trees (RRT) and its optimal variant RRT* \cite{karaman2011sampling}. Grid-based methods systematically search the environment, providing guaranteed optimal solutions under specific assumptions but suffering from computational complexity in larger or higher-dimensional spaces. Sampling-based algorithms, on the other hand, scale more effectively with dimensionality, although they typically only guarantee asymptotic optimality.

Optimization-based methods have emerged to address the limitations of classical algorithms, particularly in complex environments requiring adherence to multiple objectives and constraints simultaneously. Approaches such as Fast-Planner \cite{zhou2019robust} and EGO-Planner \cite{zhou2021ego} employ convex optimization frameworks, demonstrating significant success in cluttered and obstacle-rich scenarios. However, they rely heavily on accurate dynamic models and computationally intensive optimization processes, limiting their responsiveness and adaptability to highly dynamic conditions.

Learning-based approaches have gained popularity due to their capability to handle complex and unpredictable environments without explicit system modeling \cite{hai2025resilient, kaufmann2020deep, wu2023adaptive}. IL-based methods have demonstrated significant achievements in drone autonomous flight scenarios, where drones successfully imitate expert trajectories \cite{kaufmann2023champion, yang2023federated}. DRL techniques have similarly shown potential, especially in vision-based drone navigation tasks, effectively learning control policies directly from environmental interaction \cite{zhang2024npe, zhang2025learning}. Despite their strengths, these methods still require large-scale, task-specific datasets for training, which restricts their generalizability and robustness in new or unseen environments.

\subsection{Foundation Models}
Foundation models, characterized by extensive pre-training on massive and diverse datasets, offer powerful generalization capabilities, substantially improving robotic perception, reasoning, and adaptability across tasks \cite{bommasani2021opportunities}. Unlike traditional task-specific models, foundation models enable rapid fine-tuning to a broad spectrum of downstream tasks, reducing the dependency on extensive scenario-specific training.

Recent applications of large language models (LLMs) have demonstrated their ability to enhance robotic task planning by integrating high-level semantic understanding and logical inference capabilities. In UAV navigation, Aikins et al. \cite{aikins2024leviosa} leveraged LLMs to translate natural language instructions into actionable 3D waypoints, demonstrating effective trajectory planning from abstract human commands. Similarly, vision-language models (VLMs) have shown promise in directly interpreting visual information for UAV navigation and environmental reasoning \cite{wang2024towards}. Despite these advances, current research on applying foundation models to drone path planning primarily addresses reactive or local navigation tasks, leaving a significant gap in comprehensive benchmarking and validation of global path planning capabilities. Establishing such benchmarks is crucial for future developments, guiding researchers toward more robust and generalizable path planning solutions based on foundation models.

\section{Methodology}
\label{sec:methodology}
This section introduces the foundation model guided path planning approach. To enable spatially informed decision-making, we design a  tokenized input that combines both textual instructions and visual-perceptual information. Specifically, the prompt provided to the LLM is formulated as:

\begin{equation}
\mathcal{T}_{\text{input}} = \left[\texttt{<Prompt>}, \mathbf{P}_s, \mathbf{P}_g, \{ \mathbf{P}^n_o\}_{n=1}^{N} \right]
\end{equation}
where \(\mathbf{P}_s\) and \(\mathbf{P}_g\) denote the start and goal coordinates and \(\{ \mathbf{P}^n_o\}\) represents the position of the \(n\)-th obstacle. This structured representation is serialized into natural language tokens and fed into the foundation model to guide global path planning. We develop the FM-planner from three stages. In Stage 1, an LLM processes provided start, goal, and obstacle coordinates to generate an initial flight plan that outlines a coarse navigational trajectory. Stage 2 uses a dedicated VLM that focuses on obstacle detection and precise localization of start and goal positions directly. After the comparison between these 2 approaches, Stage 3 integrates the LLM with the fundamental visual encoder, providing the visual perception into the LLM-guided path planner. 
  
\subsection{LLM-guided Path Planner}
    The LLM-guided path planner (see Fig. \ref{fig:llmvlmframework} (a)) integrates the pretrained LLM model with prompt instruction, a waypoint parsing module, and a path refinement module. Multiple open-source pretrained LLMs are first evaluated for autonomous drone navigation tasks. The models, selected from the same microservice platform, include \textbf{Llama-3.1-8B-Instruct}, \textbf{Qwen-2.5-coder (7B and 32B variants)}, \textbf{Gemma-2 (9B and 27B variants)}, \textbf{Mistral (12B and 24B variants)}, and \textbf{Deepseek-R1}. The models are tested within identical environmental conditions using the NVIDIA Integration Manager (NIM) API, ensuring uniformity across model inference. The choice of diverse parameter counts for selected models allowed exploration of model scalability and performance trade-offs in spatial reasoning tasks. The few-shot prompting technique is applied to provide the LLM with extra context on the start point, goal, and the locations of the $2\times2\times2m$ cubic obstacles on the map. This enriched prompt allows the LLM to produce more spatially logical, collision-free waypoint sequences. The prompt is communicated to the LLM via NVIDIA’s API, yielding an initial set of waypoints. The LLM’s textual response is then parsed into numerical waypoint coordinates $\{\mathbf{P}_i\}$. Usually the generated waypoints are quite sparse. A Euclidean distance-based interpolation algorithm is adopted to smooth the path, inserting intermediate points $\{\mathbf{P}_j\}$ at $0.5 m$ intervals for smoother navigation, where $\mathbf{P}_{j|i} = \mathbf{P}_i + 0.5(\mathbf{P}_{i+1} - \mathbf{P}_i)j/\|\mathbf{P}_{i+1} - \mathbf{P}_i\|$. Regarding the waypoints around the obstacles, a safety margin $1.2m$ is maintained around the obstacle along the edges. Finally, the execution module converts the refined path into flight commands by sending position setpoints to the drone via ROS.

\begin{figure*}[!tb]
    \centering
    \includegraphics[width=\linewidth]{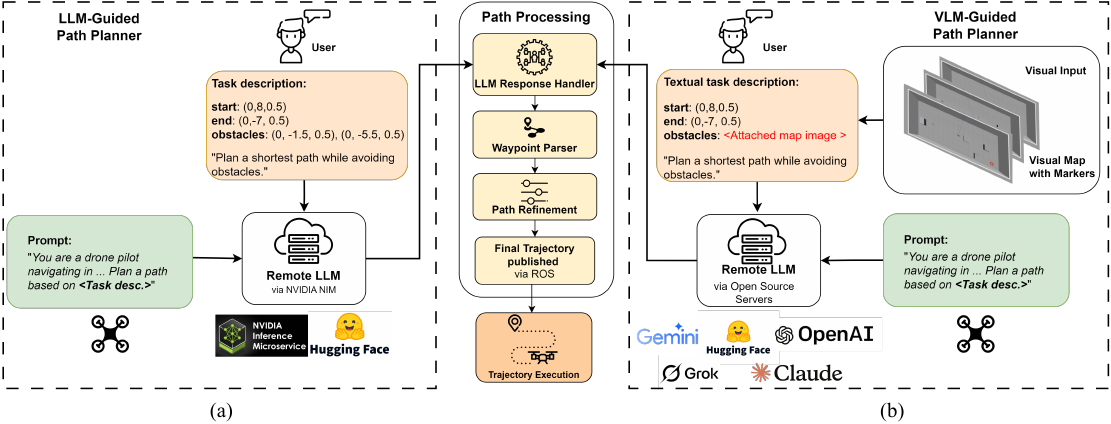}
    \caption{The framework of the foundation model guided path planners with prompt and task descriptions. (a) The LLM-guided path planner with purely textual inputs; (b) The VLM-guided path planner with textual and visual inputs. The path processing module remains the same to extract the readable waypoint list.}
    \label{fig:llmvlmframework}
\end{figure*}

\subsection{VLM-guided Path Planner}
The VLM-guided path planner explores drone navigation driven by pretrained VLMs, specifically targeting their ability to infer collision-free flight paths from visual environmental inputs (see Fig. \ref{fig:llmvlmframework} (b)). Five VLMs, \textbf{ChatGPT-4o}, \textbf{Gemini 2}, \textbf{Microsoft Copilot}, \textbf{Claude 3.5 Sonnet}, and \textbf{Grok 3}, were selected to evaluate the generative potential of multimodal models when given drone navigation scenarios in the form of annotated Gazebo simulation maps with starting coordinates $\mathbf{P}_s$ and goal coordinates $\mathbf{P}_g$. Prompt engineering techniques are applied to extract waypoint trajectories. Using few-shot prompting, each VLM first received three example cases, each consisting of a visual map paired with an obstacle-free set of $(x, y, z)$ waypoints. These examples guided the models in learning spatial patterns and generalizing safe trajectories. In an instruction-based prompting setup, models are given direct instructions (e.g., \texttt{"Analyze the visual map and generate a safe path from start to goal while avoiding obstacles"}), leveraging their multimodal capability to translate visual context into coherent navigation plans.
Each VLM returned a list of 3D waypoints representing a hypothetical collision-free trajectory from the designated start to goal positions. The raw output from the VLM is passed into the path interpolation, which applies Euclidean interpolation to smooth the motion between waypoints. A small step size of $0.1m$ is used to provide fine granularity and ensure physically feasible drone movement. This process produces a denser set of intermediate waypoints while preserving the spatial intent of the original path. The same waypoint interpolation around obstacles is adopted to smooth the path.

\begin{figure*}[!tb]
    \centering
    \includegraphics[width=\linewidth]{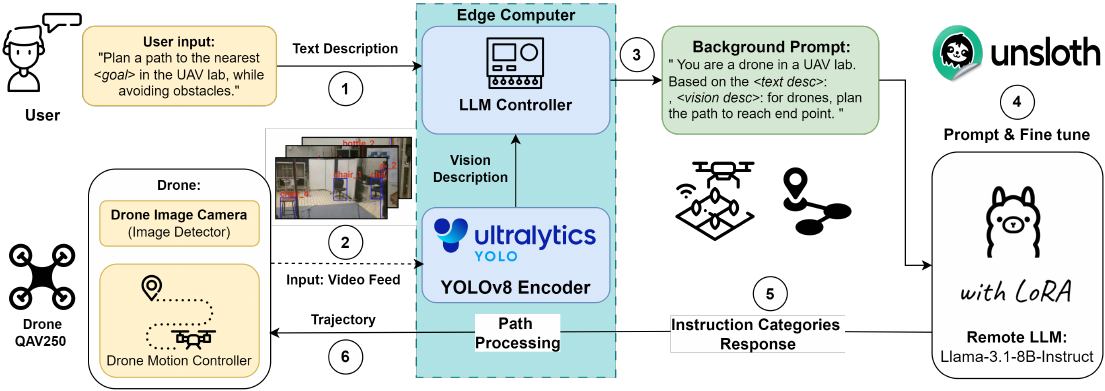}
    \caption{System architecture of the LLM-Vision-guided path planner framework. A user-provided instruction is combined with real-time visual context from the YOLOv8 vision encoder and fed into a fine-tuned LLM. The planner generates trajectories, which are executed by the drone.}
    \label{fig:llm-vision}
\end{figure*}

\subsection{LLM-guided Path Planner with Vision Encoder}
To adapt the LLM-guided path planner for practical application, the general vision encoder is integrated into the LLM module to provide the estimated positions of obstacles in real time (see Fig. \ref{fig:llm-vision}). In this work, YOLOv8 serves as the vision encoder, efficiently processing the video feed to detect and classify obstacles and relevant scene elements. Subsequently, a pretrained LLM model, fine-tuned using Low-Rank Adaptation (LoRA) via the \texttt{Unsloth.ai} platform, interprets these visual outputs to produce real-time navigation trajectories. LoRA introduces trainable low-rank updates to the frozen pre-trained weights, drastically reducing the number of trainable parameters. LoRA enables efficient adaptation by updating only a subset of the model's parameters, specifically within the attention heads and feed-forward layers. 
Let \(\mathbf{W}_0 \in \mathbb{R}^{d \times k}\) denote a weight matrix in the original LLM. Instead of updating \(\mathbf{W}_0\) directly, LoRA learns a low-rank decomposition:
\begin{equation}
\Delta \mathbf{W} = \mathbf{B} \mathbf{A},
\end{equation}
where \(\mathbf{B} \in \mathbb{R}^{d \times r}\) and \(\mathbf{A} \in \mathbb{R}^{r \times k}\) with \(r \ll \min(d, k)\). The updated forward computation becomes:
\begin{equation}
\mathbf{y} = \mathbf{W}_0 \mathbf{x} + \Delta \mathbf{W} \mathbf{x} = \mathbf{W}_0 \mathbf{x} + \mathbf{B}(\mathbf{A} \mathbf{x}),
\end{equation}
where \(\mathbf{x} \in \mathbb{R}^{k}\) is the input and \(\mathbf{y} \in \mathbb{R}^{d}\) is the output.
This approach keeps \(\mathbf{W}_0\) frozen and only updates \(\mathbf{A}\) and \(\mathbf{B}\), achieving parameter-efficient adaptation with \(\mathcal{O}(dr + rk)\) additional parameters instead of \(\mathcal{O}(dk)\). This efficiency is especially valuable for deployment on resource-constrained hardware, such as edge computers used in autonomous drones. The training objective is to minimize the cross-entropy loss between the predicted waypoint output $\hat{\mathbf{P}}$ and the ground truth waypoints \({\mathbf{P}}\) over a dataset of \(N\) task samples:
\begin{equation}
\mathcal{L}_{\text{CE}} = - \frac{1}{N} \sum_{i=1}^{N} \sum_{j=1}^{C} \mathbf{P}^{(i)}_j \log \hat{\mathbf{P}}^{(i)}_j,
\end{equation}
where \(C\) is the number of waypoints and \(\mathbf{P}^{(i)}_j\) is the ground truth waypoints \(j\) in task \(i \in N\).

Domain-specific datasets from simulation and real-world drone navigation scenarios are used to fine-tune the LLM, refining its natural language understanding and spatial reasoning for navigation tasks. Simultaneously, tailored prompts are developed with explicit drone parameters to guide the model toward generating structured, accurate waypoint sequences. A synthetic dataset is generated using a custom Python script to cover a wide range of drone navigation scenarios. Each entry included coordinates for a start point, goal, and multiple obstacle positions. For each scenario, waypoints are algorithmically computed to guarantee at least a $1m$ clearance from every obstacle, ensuring collision-free navigation paths.

The vision encoder identifies environmental features, outputting structured data including class labels marked by two-dimensional (2D) bounding boxes. The depth information is further provided to estimate the 3D position. These outputs are tokenized and combined with a carefully constructed domain-specific prompt to inform navigation in complex indoor environments. The combined visual-context input is then fed into the LLM in a zero-shot prompting technique, where the model infers navigation paths without direct prior examples. Based solely on visual labels and provided instructions, the LLM generates a sequence of waypoints in a two-dimensional plane (maintaining a fixed altitude) for the navigation trajectory.

\section{Experiments and Results}
\label{sec:results}

\subsection{Simulation Setup}
These experiments aimed to evaluate how accurately and efficiently pretrained foundation models can plan autonomous drone paths using prompt instructions. Experiments were conducted within a constrained indoor Gazebo simulation environment, simulating realistic drone operational scenarios. The drone (controlled via ROS Noetic and MAVROS) navigated a 3D space populated with static cubic obstacles (size of $2m \times 2m \times 2m$). The drone is the default Iris model with the PX4 controller. The LLMs produced waypoint lists with prompts specifying start position $\mathbf{P}_s$, goal position $\mathbf{P}_g$, and obstacle coordinates $\{\mathbf{P}_o\}$. The raw waypoints were then parsed and refined to ensure smooth trajectories, with intermediate waypoints interpolated using a Euclidean distance-based algorithm with a step size of $0.5m$. The simulations were running on a workstation with an RTX 3090 GPU (24 GB of G6X memory).

\begin{table*}[!bt]
\centering
\rowcolors{2}{gray!20}{white}
\small
\caption{Comparative evaluation of LLM models over various obstacle avoidance scenarios}
\label{tab:llm-2v3-obstacles}
\begin{tabular}{l
                cccc
                cccc}
\toprule
& \multicolumn{4}{c}{\bfseries 2 Obstacles} 
& \multicolumn{4}{c}{\bfseries 3 Obstacles} \\
\cmidrule(r){2-5} \cmidrule(l){6-9}
\bfseries Model 
& \parbox[c]{1.5cm}{\centering ACT (s) $\downarrow$} 
& \parbox[c]{1.5cm}{\centering PL (m) $\downarrow$} 
& SR $\uparrow$ 
& ESS $\uparrow$ 
& \parbox[c]{1.5cm}{\centering ACT (s) $\downarrow$} 
& \parbox[c]{1.5cm}{\centering PL (m) $\downarrow$} 
& SR $\uparrow$
& ESS $\uparrow$\\
\midrule
Llama-3.1-8B-Instruct 
  & 80s & 14.3 & 100\% & \textbf{1.250} 
  & 90s & 16.1 & 100\% & \textbf{1.111} \\

Deepseek‑R1                    
  & –         & –  &   0\%  &   0.000  
  & –         &  – &   0\%  &   0.000    \\

Qwen‑2.5‑coder‑7B     
  & 90s &  16.1 &  90\%  & 1.000 
  & 102s & 18.2 & 100\%  & 0.980 \\

Qwen‑2.5‑coder‑32B   
  & 86s &  15.4 &  90\%  & 1.046 
  & 87s &  15.5 &  80\%  & 0.920 \\

Gemma‑2‑9B‑it                  
  & 88s &  15.7 &  70\%  & 0.795 
  & 90s &  16.1 &  60\%  & 0.667 \\

Gemma‑2‑27B‑it                 
  & 88s &  15.7 &  80\%  & 0.909 
  & 96s &  17.2 &  80\%  & 0.833 \\

Mistral‑nemo‑12B     
  & 83s &  14.8 &  50\%  & 0.602 
  & 105s &  18.8 &  60\%  & 0.571 \\

Mistral‑small‑24B     
  & 83s &  14.8 &  70\%  & 0.843 
  & 99s &  17.7 &  70\%  & 0.707 \\
\midrule
A* \cite{zhou2022uav}    
  & 80s &  14.3 &  100\%  & \textbf{1.250} 
  & 87s &  15.6 &  100\%  & \textbf{1.115} \\

RRT \cite{hu2021efficient}    
  & 84s &  15.0 &  50\%  & 0.595 
  & 84s &  15.0 &  90\%  & 1.071 \\

Q-learning \cite{wu2023adaptive}    
  & 80s &  14.3 &  50\%  & 0.625 
  & 87s &  15.6 &  100\%  & \textbf{1.115} \\   

\bottomrule
\end{tabular}
\end{table*}

\begin{table*}[!tb]
\centering
\small
\caption{Detailed LLM Model Comparison (Llama‑3.1-8B-Instruct vs Qwen‑2.5‑coder‑7B)}
\label{tab:llm-llama-vs-qwen}
\renewcommand{\arraystretch}{1.2}
\begin{tabular}{>{}l
                c
                p{4.2cm}
                c
                c
                c
                c}
\toprule
\bfseries Model 
  & \bfseries \# Obs. 
  & \bfseries Coordinates of Obstacles 
  & \bfseries ACT (s) $\downarrow$
  & \bfseries PL (m) $\downarrow$
  & \bfseries SR $\uparrow$
  & \bfseries ESS $\uparrow$ \\
\midrule
\multirow{4}{*}{Llama‑3.1-8B-Instruct} 

  & \cellcolor{gray!20}2
    & \cellcolor{gray!20}\shortstack[l]{
      (0,\,-1.5,\,0.5);
      (0,\,-5.5,\,0.5)
    }
  & \cellcolor{gray!20}80s
  & \cellcolor{gray!20}14.3
  & \cellcolor{gray!20}100\%
  & \cellcolor{gray!20}1.250 \\

  & 3
    & \shortstack[l]{
      (0,\,-1.5,\,0.5);
      (0,\,-5.5,\,0.5);\\
      (-2.5,\,-3.5,\,0.5)
    }
  & 90s
  & 16.1
  & 100\%
  & 1.111 \\

   & \cellcolor{gray!20}4
    & \cellcolor{gray!20}\shortstack[l]{
      (0,\,-1.5,\,0.5);
      (0,\,-5.5,\,0.5);\\
      (-2.5,\,-3.5,\,0.5);
      (-2.5,\,0.5,\,0.5)
    }
  & \cellcolor{gray!20}100s
  & \cellcolor{gray!20}17.9
  & \cellcolor{gray!20}90\%
  & \cellcolor{gray!20}0.900 \\

  & 5
    & \shortstack[l]{
      (0,\,-1.5,\,0.5);
      (0,\,-5.5,\,0.5);\\
      (-2.5,\,-3.5,\,0.5);
      (-2.5,\,0.5,\,0.5);\\
      (-1.5,\,3.5,\,0.5)
    }
  & 110s
  & 19.7
  & 90\%
  & 0.818 \\

\midrule
\multirow{4}{*}{Qwen‑2.5‑coder‑7B}
  & \cellcolor{gray!20}2
    & \cellcolor{gray!20}\shortstack[l]{
      (0,\,-1.5,\,0.5);
      (0,\,-5.5,\,0.5)
    }
  & \cellcolor{gray!20}90s
  & \cellcolor{gray!20}16.1
  & \cellcolor{gray!20}90\%
  & \cellcolor{gray!20}1.000 \\

  & 3
    & \shortstack[l]{
      (0,\,-1.5,\,0.5);
      (0,\,-5.5,\,0.5);\\
      (-2.5,\,-3.5,\,0.5)
    }
  & 102s
  & 18.2
  & 100\%
  & 0.980 \\

  & \cellcolor{gray!20}4
    & \cellcolor{gray!20}\shortstack[l]{
      (0,\,-1.5,\,0.5);
      (0,\,-5.5,\,0.5);\\
      (-2.5,\,-3.5,\,0.5);
      (-2.5,\,0.5,\,0.5)
    }
  & \cellcolor{gray!20}110s
  & \cellcolor{gray!20}19.7
  & \cellcolor{gray!20}90\%
  & \cellcolor{gray!20}0.818 \\

  & 5
    & \shortstack[l]{
      (0,\,-1.5,\,0.5);
      (0,\,-5.5,\,0.5);\\
      (-2.5,\,-3.5,\,0.5);
      (-2.5,\,0.5,\,0.5);\\
      (-1.5,\,3.5,\,0.5)
    }
  & – 
  & 0/10
  & 0\%
  & 0.000 \\

\bottomrule
\end{tabular}
\end{table*}

\subsection{Evaluation Metrics}
To evaluate the effectiveness of each pretrained language model in drone path planning, three primary metrics were used: Success Rate (SR), Average Completion Time (ACT), Path Length (PL) and the derived Efficiency-Success Score (ESS). SR is the percentage of trials where the drone reached the goal without collision, indicating reliability and obstacle avoidance capability. ACT quantifies the average time taken to complete a successful navigation task, reflecting operational efficiency and the path smoothness. Considering these metrics in isolation can be misleading; for example, a model might finish quickly but have a low success rate or achieve high success but take too long. To address this, we introduce the ESS as a composite metric, defined as $ESS = SR/ACT\, (\%/sec)$.

\subsection{Baseline Planners}
We selected three typical path planners as the baselines, namely the A* planner \cite{zhou2022uav}, the RRT planner \cite{hu2021efficient} and the Q-learning planner \cite{wu2023adaptive}. To maintain the same environment settings, we discretize the workspace 
\(
\mathcal{X} = \{(x,y)\mid x\in[-3,3],\,y\in[-8,8]\}\subset\mathbb{R}^2
\)
into a uniform 2D $50\times50$ grid. Denote the grid-cell size by
\(
\Delta_x = \Delta_y = 0.12m.
\)
Start and goal positions are $\mathbf P_s$ and $\mathbf P_g$, and obstacle centers $\{\mathbf P_o\}$ are expanded into $1m\times1m$ squares in the grid.

\subsubsection{A* Planner} A* is a graph-search algorithm that finds an optimal path on a discretized grid by combining the actual cost from the start node with a heuristic estimate to the goal. We construct a 4‐connected grid graph $G=(V,E)$, where each vertex $v\in V$ corresponds to a free cell.  For an edge $(u,v)\in E$, the cost is
\(
c(u,v) = \Delta_x =  0.12 m.
\)
Define the objective function and heuristic functions as follows:
\[
g(n) = \min_{\text{path }s\to n}\sum_{\!(u,v)\!}c(u,v),\quad
h(n) = |x_n - x_g| + |y_n - y_g|,
\]
and the A* total cost
\(
f(n) = g(n) + h(n).
\)
With admissible Manhattan heuristic $h(n)$, A* optimize the node connection and returns an optimal path.

\subsubsection{RRT Planner} RRT is a sampling-based planner that incrementally builds a search tree rooted at the start position by repeatedly sampling random states and extending the nearest tree node toward each sample by a fixed step size.
We grow a tree $T$ as:
\(
x_{\mathrm{rand}}\sim \mathrm{Uniform}(\mathcal X_{\mathrm{free}})
\),
\(
x_{\mathrm{near}} = \arg\min_{x\in T}\|x - x_{\mathrm{rand}}\|
\), and the update rule is 
\begin{equation}
    x_{\mathrm{new}} = x_{\mathrm{near}} + \delta\frac{x_{\mathrm{rand}} - x_{\mathrm{near}}}{\|x_{\mathrm{rand}} - x_{\mathrm{near}}\|},
\end{equation}
where $\delta=1$\,cell.  If $x_{\mathrm{new}}\in\mathcal X_{\mathrm{free}}$, add it to $T$.  Stop when 
\(
\|x_{\mathrm{new}} - \mathbf P_g\| \le \epsilon,\quad \epsilon = 2\ \text{cells}\,.
\)
RRT is probabilistically complete but not guaranteed optimal.

\subsubsection{Q-learning Planner} We implement a simple greedy policy that, at every step, selects the neighboring grid cell minimizing Euclidean distance to the goal.
At each discrete time step $k$, with current cell $x_k$, choose the next state 
\begin{equation}
 x_{k+1} = \arg\min_{x\in \mathcal{N}_4(x_k),\,x\notin\mathcal O}
\|x - \mathbf P_g\|,
\end{equation}
where $\mathcal{N}_4(x_k)$ are the 4‐neighbors and $\mathcal O$ is the obstacle set. If no neighbor reduces the distance, pick any adjacent free cell. We limit the horizon with $K_{\max}=500$ steps. This myopic policy requires no training and highlights the benefits of full DRL agents with long-horizon value estimation.

\subsection{Benchmarking LLM-guided Path Planner}

\subsubsection{Base Comparison} We first evaluated the performance and generalizability of several LLM models and baseline planners across two distinct scenarios involving different obstacle complexities: a scenario ($S1$) with two obstacles and another scenario ($S2$) with three obstacles (see Fig. \ref{fig:vlm-test}). Specifically, the scenario $S1$ was configured with start and goal positions at $\mathbf{P}_s = [0.0, 6.0, 0.5]$ and $\mathbf{P}_g = [0.0, -7.0, 0.5]$, respectively, with obstacle locations defined as $\{\mathbf{P}_o\} = \{[0.0, -1.5, 0.5], [0.0, -5.5, 0.5]\}$. Scenario $S2$, featuring an additional obstacle, used $\mathbf{P}_s = [0.0, 6.0, 0.5]$, $\mathbf{P}_g = [-2.5, -7.0, 0.5]$, and obstacle coordinates $\{\mathbf{P}_o\} = \{[0.0, -1.5, 0.5], [0.0, -5.5, 0.5], [-2.5, -3.5, 0.5]\}$. Their comparative performances are summarized in Table~\ref{tab:llm-2v3-obstacles}.

In both scenarios, Llama-3.1-8B-Instruct matches the optimal A* planner on reliability and efficiency, and significantly outperforms all other neural planners. In the two-obstacle scenario (S1), Llama-3.1-8B-Instruct achieves a 100\% success rate (SR) in average $80 s$ over a $14.3 m$ trajectory (ESS = 1.250). By contrast, DeepSeek-R1 fails in all 10 trials, underscoring the limitations of purely reactive methods. The mid-sized Qwen-2.5-7B variant attains a 90\% SR with longer paths and lower ESS, reflecting the trade-off between distance and navigation precision.

When complexity increases with a third obstacle (S2), Llama-3.1-8B-Instruct again sustains 100\% SR in average $90 s$ over a $16.1 m$ path (ESS = 1.111). Similarly, Qwen-2.5-7B attains 100\% SR with ESS = 0.980, demonstrating strong generalization. In contrast, the Qwen-2.5-32B and both Gemma variants see their SR drop to approximately 80\%, highlighting scalability challenges in more complex environments. The RRT planner is able to provide the shortest path but is not stable due to its random search mechanism. Given their consistently superior SR and ESS, Llama-3.1-8B-Instruct and Qwen-2.5-7B were selected for further detailed comparison.

\begin{table}[!tb]
\centering
\small
\caption{Effect of Speed on Navigation Success for Llama‑3.1-8B-Instruct}
\label{tab:llm-speed-comparison}
\renewcommand{\arraystretch}{1.2}
\begin{tabular}{@{}lccccc@{}}
\toprule
\bfseries Scenario 
 & \bfseries Rate 
 & \bfseries ACT $\downarrow$
 & \bfseries Speed $\uparrow$
 & \bfseries SR $\uparrow$
  & \bfseries ESS $\uparrow$\\
\midrule
\multirow{5}{*}{\itshape 2 Obstacles} 
 & 20Hz  & 80s  & 0.179m/s & 100\% & 1.250\\
 & 30Hz  & 75s  & 0.191m/s & 100\% & 1.333\\
 & 40Hz  & 70s  & 0.204m/s & 100\% & 1.429\\
 & 50Hz  & 60s  &  0.238m/s &  20\% & 0.333\\
 & 60HZ  & –           &  – &   0\% & 0.000\\
\midrule
\multirow{5}{*}{\itshape 3 Obstacles} 
 & 20Hz  & 90s  & 0.179m/s & 100\% & 1.111\\
 & 30Hz  & 80s  & 0.201m/s & 100\% & 1.250\\
 & 40Hz  & 65s  &  0.248m/s &  90\% & 1.385\\
 & 50Hz  & 60s  &  0.268m/s
 &  90\% & 1.500\\
 & 60Hz  & –  &  – &  0\% & 0.000\\
\bottomrule
\end{tabular}
\end{table}

\subsubsection{Detailed Comparison}
The detailed comparative evaluations are conducted between Llama‑3.1-8B-Instruct and Qwen‑2.5-coder-7B under varying levels of obstacle complexity. Specifically, the start and goal positions were consistently set to $\mathbf{P}_s = [0.0, 6.0, 0.5]$ and $\mathbf{P}_g = [-2.5, -7.0, 0.5]$, respectively. The obstacle configurations and comparative results are presented in Table~\ref{tab:llm-llama-vs-qwen}. Across all tested scenarios, Llama‑3.1-8B-Instruct consistently outperformed Qwen‑2.5 in terms of both reliability and navigation accuracy. While both models initially achieved high success rates with two and three obstacles, the performance of Qwen‑2.5 degraded significantly as the obstacle count increased, culminating in a complete failure in scenarios with five obstacles. Meanwhile, the Qwen‑2.5 generates a longer PL to ensure safety. In contrast, Llama‑3.1-8B-Instruct maintained a robust success rate $\geq 90\%$ and consistently accurate navigation even in the most challenging scenario involving five obstacles. These findings highlight Llama‑3.1-8B-Instruct's strong generalization capability and robustness, supporting its selection for subsequent operational speed testing.

\begin{figure}[!tb]
    \centering
    \includegraphics[width=\linewidth]{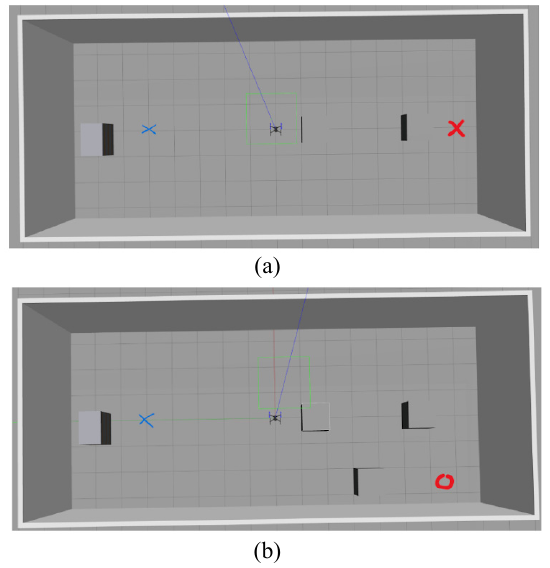}
    \caption{Input bird's-eye view images for the VLM-guided path planner test. (a) Two obstacles; (b) three obstacles.}
    \label{fig:vlm-test}
\end{figure}

\begin{figure}[!tb]
    \centering
    \includegraphics[width=\linewidth]{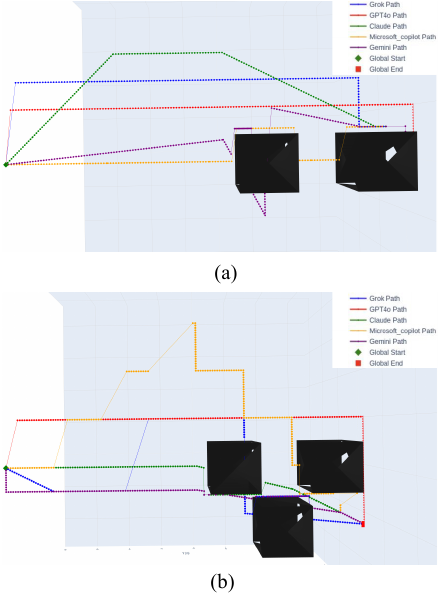}
    \caption{Paths generated from various VLMs. (a) Mission with two obstacles; (b) Mission with three obstacles. The VLMs rarely generate  optimal paths.}
    \label{fig:vlm-traj}
\end{figure}

\subsubsection{Speed Test}
To further evaluate the reliability and practical safety of the LLM-guided path planner, we assessed the performance of Llama‑3.1-8B-Instruct under varying ROS publishing rates, reflecting different operational flight speeds. The start and goal positions were fixed at $\mathbf{P}_s = [0.0, 6.0, 0.5]$ and $\mathbf{P}_g = [-2.5, -7.0, 0.5]$. The environments included a two-obstacle scenario ($\mathbf{P}_o \!= \!\{[0.0, -1.5, 0.5], [0.0, -5.5, 0.5]\}$) and a three-obstacle scenario ($\mathbf{P}_o=\{[0.0, -1.5, 0.5], [0.0, -5.5, 0.5], [-2.5, -3.5, 0.5]\}$). Test outcomes are summarized in Table~\ref{tab:llm-speed-comparison}. Results indicate that the Llama‑3.1-8B-Instruct model reliably maintained a perfect 100\% success rate at ROS publishing rates up to 40 Hz. However, performance rapidly deteriorated at 50 Hz and higher rates due to insufficient re-planning intervals between successive control updates. Consequently, a maximum practical control frequency of approximately 40 Hz is recommended, balancing responsiveness with reliable navigation performance.

\begin{table}[!tb]
  \centering
  \small
  \caption{Llama-3.1-8B-Instruct Fine-tuning Parameters}
  \label{tab:fine-tuning-params}
  \renewcommand{\arraystretch}{1.2}
  \begin{tabular}{@{}lc@{}}
    \toprule
    \bfseries Hyperparameter          & \bfseries Value                      \\
    \midrule
    Batch Size                        & 4                                    \\
    Epochs                            & 60                                   \\
    Learning Rate                     & $2\times10^{-4}$                     \\
    Optimiser                         & AdamW (8-bit)                        \\
    Activation Function               & ReLU (MLP), Swish (LLM default)      \\
    Weight Decay                      & 0.01                                 \\
    \bottomrule
  \end{tabular}
\end{table}

\begin{figure}[!tb]
    \centering
    \includegraphics[width=\linewidth]{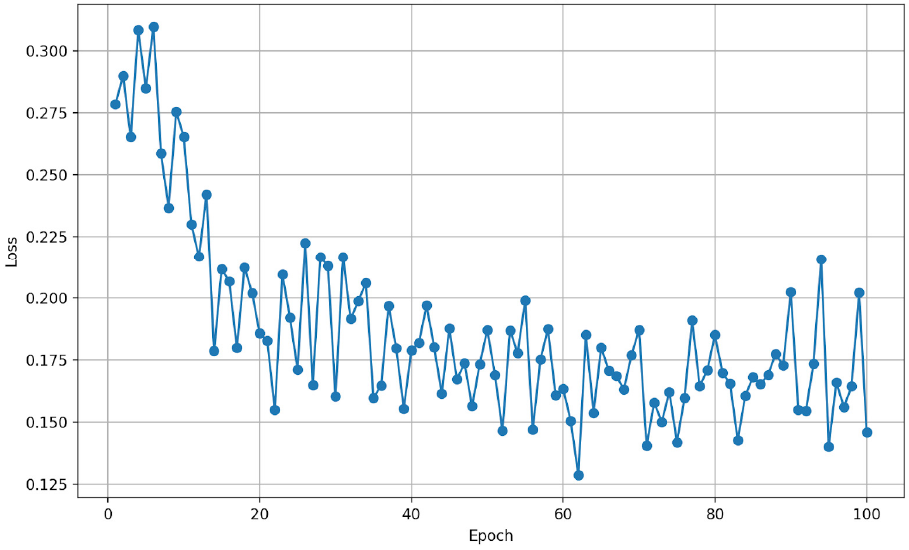}
    \caption{The training curve of Llama-3.1-8B-Instruct during fine-tuning.}
    \label{fig:finetunning}
\end{figure}

\subsection{Benchmarking VLM-guided Path Planner}
\subsubsection{Settings}
This part evaluates the effectiveness of VLM-guided path planner using high-resolution visual map inputs instead of purely textual coordinate-based inputs. In contrast to the LLM-guided path planner approach, VLM-guided methods directly leverage visual context obtained from environmental imagery, emphasizing spatial reasoning grounded in visual cues. Experiments were conducted within the Gazebo ROS Noetic simulation environment, where the drone navigated from clearly defined starting positions to goal positions while avoiding cubic obstacles of size $1m \times 1m \times 1m$ placed throughout the simulated space. Precise starting and goal 3D coordinates were parsed and provided as inputs to the respective VLM models. Several representative VLMs, including ChatGPT-4o, Gemini 2, Microsoft Copilot, Claude 3.5 Sonnet, and Grok 3, were systematically evaluated to assess their potential for purely vision-based spatial reasoning. Fig.~\ref{fig:vlm-test} illustrates the annotated visual maps utilized as inputs for the VLMs, clearly marking the drone’s initial (blue) and target (red) positions.

\subsubsection{Analysis}
Fig. \ref{fig:vlm-traj} presents trajectories generated by all five VLMs within each scenario. GPT-4o and Claude produced relatively coherent and feasible paths, successfully avoiding obstacles, though their trajectories were often nonoptimal in terms of distance or smoothness. Microsoft Copilot exhibited moderate performance, generating valid but overly conservative paths. In contrast, Gemini and Grok struggled to produce reliable trajectories, frequently exhibiting collisions or erratic behavior.
Compared to the LLM-based approach, the VLM-guided planners were less consistent in generating safe and efficient paths. Despite having access to rich visual information, their performance was limited by a lack of robust spatial reasoning and insufficient generalization. On the other hand, Llama-3.1-8B-Instruct consistently delivered reliable results using only textual spatial inputs, demonstrating superior logical inference and planning capabilities. To combine the strengths of both modalities, we developed an LLM-guided path planner enhanced with a vision encoder, enabling real-time visual perception while maintaining strong spatial reasoning.

\begin{figure*}[!tb]
    \centering
    \includegraphics[width=\linewidth]{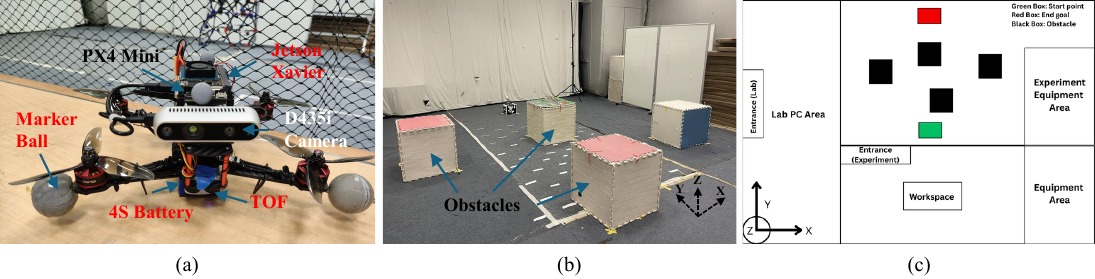}
    \caption{The configuration of physical experiments. (a) The configuration of QAV drone with the companion computer; (b) The autonomous flight area with multiple obstacles; (c) The layout of the flight area.}
    \label{fig:phycfg}
\end{figure*}

\begin{figure*}[!tb]
    \centering
    \includegraphics[width=\linewidth]{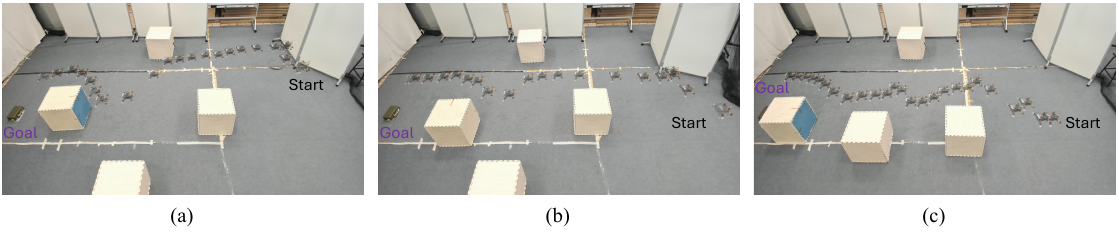}
    \caption{The real-world autonomous flight with the developed LLM-Vision planner under various tasks in Table \ref{tab:uav-navigation-results}. (a) Task 1; (b) Task 2; (c) Task 3.}
    \label{fig:phytask}
\end{figure*}

\subsection{Fine-tuning for LLM-Vision Planner}
With Llama-3.1-8B-Instruct selected as the path planner module, we further fine-tune the model with LoRA. The hyperparameters were selected based on empirical testing and prior studies on efficient LLM adaptation (see Table~\ref{tab:fine-tuning-params}). To support this process, we curated domain-specific datasets derived from both simulated and real-world drone navigation scenarios. In parallel, we designed task-specific prompts incorporating explicit drone parameters to guide the model in generating structured and accurate waypoint sequences. A synthetic dataset comprising 5,000 entries was generated to represent diverse navigation scenarios. Each entry includes coordinates for a start point $\mathbf{P}_s$, a goal point $\mathbf{P}_g$, and multiple obstacle positions $<\mathbf{P}^i_o>$. Since the RRT planner provides optimal shortest paths, waypoints were computed using the RRT planner to maintain a minimum $1m$ clearance from all obstacles, ensuring safe and collision-free trajectories.

The model was trained over 100 epochs, and the training performance was monitored using cross-entropy loss, which quantifies how closely the model's predicted distributions align with the ground truth. As shown in Fig.~\ref{fig:finetunning}, the training loss converged to approximately 0.15 after 60 epochs, indicating stable and effective fine-tuning.

\begin{table}[!tb]
  \centering
  \small
  \caption{LLM-Vision Planner Tasks and Results}
  \label{tab:uav-navigation-results}
  \renewcommand{\arraystretch}{1.2}
  \begin{tabular}{c  c  c  c  c c}
    \toprule
      \bfseries $\mathbf{P}_s/\textcolor{red}{\mathbf{P}_g}$ 
      & \bfseries \#Obs 
      & \bfseries $\{\mathbf{P}_o\}$ 
      & \bfseries ACT 
      & \bfseries Success & RT\\
    \midrule
      \cellcolor{gray!20}\shortstack[l]{ Task1: \\ (0.5,\,1.0,\,0.6)\\  \textcolor{red}{ (0.5,\,6.3,\,0.6)}}
      & \cellcolor{gray!20}4
      & \cellcolor{gray!20}\shortstack[c]{
        (0.5,\,2.3,\,0.6)\\
        (-1.2,\,3.3,\,0.6)\\
        (2.5,\,4.2,\,0.6)\\
        (0.5,\,4.4,\,0.6)
        }
      & \cellcolor{gray!20}25s
      & \cellcolor{gray!20}Yes & \cellcolor{gray!20}9.6s \\

 \shortstack[l]{ Task2: \\(-0.5,\,0.5,\,0.6)\\    \textcolor{red}{(0.5,\,6.3,\,0.6)}}
      & 4
      & \shortstack[c]{
        (0.5,\,2.3,\,0.6)\\
        (-1.2,\,3.3,\,0.6)\\
        (2.5,\,4.2,\,0.6)\\
        (0.0,\,4.4,\,0.6)
        }
      & 23s
      & Yes & 9.8s\\

 \cellcolor{gray!20}\shortstack[l]{Task3: \\ (-0.5,\,1.0,\,0.6)\\  \textcolor{red}{(0.5,\,6.3,\,0.6)}}
      & \cellcolor{gray!20}4
      & \cellcolor{gray!20}\shortstack[c]{
        (0.0,\,2.3,\,0.6)\\
        (-0.5,\,3.3,\,0.6)\\
        (2.5,\,4.2,\,0.6)\\
        (0.0,\,5.0,\,0.6)
        }
      & \cellcolor{gray!20}20s
      & \cellcolor{gray!20}Yes & \cellcolor{gray!20}9.4s\\
    \bottomrule
  \end{tabular}
\end{table}

\subsection{Physical Experiments}

To validate the effectiveness of the proposed LLM-Vision-guided path planner, we deployed the fine-tuned model on a QAV250 racing drone equipped with a Jetson Xavier NX edge computer. Physical experiments were conducted in an indoor flight space of size $8 \times 10 \times 5m$, as illustrated in Fig.~\ref{fig:phycfg}. The drone operated in offboard mode, receiving self-localization data streamed at 120 Hz from an OptiTrack motion capture system. The start and goal positions were manually specified by the user, while obstacle information was extracted from RGB and depth images using the onboard vision encoder. When the LLM-Vision-guided planner was employed, real-time perception was provided by the RealSense D435i RGBD camera. The captured images were resized and processed with the vision encoder before being passed to the Llama-3.1-8B-Instruct model, which was requested from the Hugging Face server and executed at each planning step (see Fig.~\ref{fig:llm-vision}, Step 2).

We evaluated the planner across multiple scenarios with varying start positions and obstacle configurations. The test setup and performance results are summarized in Table~\ref{tab:uav-navigation-results}, and example flight trajectories are shown in Fig.~\ref{fig:phytask}. More results are available in the supplementary video. Across all tested conditions, the LLM-Vision-guided planner successfully generated obstacle-avoiding trajectories, guiding the drone to its targeted goal. As illustrated in Fig.~\ref{fig:phytask} (c), when obstacles were placed closely along the direction to the goal, the planner was still able to generate aggressive waypoints along the obstacles. The model’s average reasoning time (RT) was approximately 9 seconds per query, which is acceptable for real-time, onboard autonomous flight applications since the global path is not required to update at high frequency.

\section{Conclusion}
\label{sec:conclusion}
In this work, we introduced a novel foundation model-guided framework named FM-planner for drone path planning, systematically benchmarking multiple LLM and VLM approaches. Our extensive evaluation revealed that pretrained foundation models, particularly VLMs, lack global spatial reasoning capabilities compared to LLMs without specific textual descriptions. In particular, fine-tuned LLMs integrated with vision encoders demonstrate robust spatial reasoning and real-time obstacle awareness, making them well-suited for practical drone global planning tasks. Physical experiments confirmed the feasibility and reliability of our approach in real-world autonomous flight. Future work includes exploring advanced multimodal foundation models for complex, dynamic environments and further optimizing real-time computational performance for broader operational scenarios.

\bibliographystyle{IEEEtran}
\bibliography{IEEEabrv,xiao}

\vfill
\end{document}